\documentclass{article}
\usepackage[letterpaper, margin=1in]{geometry} % for wide text
\usepackage{graphicx} % Required for inserting images
\usepackage{subcaption}
\usepackage{hyperref}
\usepackage{amssymb}
\usepackage{bm}
\usepackage{amsmath}
\usepackage{mathtools}
\usepackage[ruled,linesnumbered]{algorithm2e}
\usepackage{multirow}
\usepackage{xcolor}
\usepackage{lscape}
\usepackage{enumitem}
\usepackage{subcaption}
\usepackage{listings}
\usepackage{fancyvrb}

\def\F{\mathcal{F}}

\def\T{\mathcal{T}}

\title{PDE Generalization of In-Context Operator Networks:\\ A Study on 1D Scalar Nonlinear Conservation Laws}
\author{Liu Yang, Stanley J. Osher\footnote{Corresponding Author: sjo@math.ucla.edu. All code will be deposited to \url{https://github.com/LiuYangMage/in-context-operator-networks}}}
\date{Department of Mathematics, UCLA, Los Angeles, CA 90095, USA}

\begin{document}

\maketitle

\begin{abstract}
Can we build a single large model for a wide range of PDE-related scientific learning tasks? Can this model generalize to new PDEs, even of new forms, without any fine-tuning? In-context operator learning and the corresponding model In-Context Operator Networks (ICON) represent an initial exploration of these questions. The capability of ICON regarding the first question has been demonstrated previously. In this paper, we present a detailed methodology for solving PDE problems with ICON, and show how a single ICON model can make forward and reverse predictions for different equations with different strides, provided with appropriately designed data prompts. We show the positive evidence to the second question, i.e., ICON can generalize well to some PDEs with new forms without any fine-tuning. This is exemplified through a study on 1D scalar nonlinear conservation laws, a family of PDEs with temporal evolution. We also show how to broaden the range of problems that an ICON model can address, by transforming functions and equations to ICON's capability scope. We believe that the progress in this paper is a significant step towards the goal of training a foundation model for PDE-related tasks under the in-context operator learning framework.
\end{abstract}

\section{Introduction}

Looking back to the evolution of neural network solvers for partial differential equations (PDEs), we see a three-act progression. 

Act 1 focuses on approximating the solution functions with neural networks. Typical (but not exhaustive) examples include~\cite{lagaris1998artificial} which represents an early exploration to solve PDEs with neural networks, \cite{han2017deep} and \cite{Han2018Solving} for high-dimensional parabolic PDEs and backward stochastic differential equations, Deep Galerkin Method (DGM)~\cite{Sirignano2018DGM}, Deep Ritz Method (DRM)~\cite{E2018deepRitz}, Physics-Informed Neural Networks (PINNs)~\cite{Raissi2019PINN}, Weak Adversarial Network (WAN)~\cite{Zang2020Weakadversarial} to impose PDEs of different forms in the training loss function, \cite{yang2020potential,Ruthotto2020machine} to solve high-dimensional optimal transport and mean-field control/game problems, and APAC-Net~\cite{Lin2021Alternating} for mean-field game problems with the primal-dual formulation.

Despite their success, in Act 1 the neural network needs to be trained again when the solution function changes due to changes in the equation or the initial/boundary conditions. Such limitation leads to Act 2, where the neural networks are employed to approximate solution operators, namely ``operator learning''. Here an operator transforms one or multiple input functions, termed as the ``condition'' in this paper, to an output function, termed as the ``quantity of interest (QoI)'' in this paper. A wide range of scientific machine learning tasks can be formulated as operator learning problems. Take the task of solving time-independent PDEs for instance, the condition could be the coefficient function, with QoI being the solution, and different equations corresponding to different operators. For optimal control problems, the condition could correspond to the initial state, while the QoI embodies the control signal, with different control dynamics corresponding to different operators. 

Operator learning can be traced back to~\cite{chen1995universal, chen1995approximation} where shallow neural networks are used to approximate nonlinear operators. More recent examples include (but are not limited to) \cite{khoo2021solving} for parametric PDE problems, \cite{zhu2018bayesian} for problems governed by stochastic PDEs, PDE-Net~\cite{long2018pde} for PDEs with temporal evolution, Fourier Neural Operator (FNO)~\cite{li2021fourier,kovachki2023neural} with integral kernel in Fourier space to learn the solution operator, Deep Operator Network (DeepONet)~\cite{lu2021learning,wang2021learning} which maps the parameters or the initial/boundary conditions to the solutions, \cite{bhattacharya2021model} and Physics-Informed Neural Operators (PINO)~\cite{li2021physics} for parametric PDEs, \cite{sanchez2020learning,pfaff2020learning,brandstetter2021message} with graph neural networks.
Other related work includes~\cite{kochkov2021machine,kissas2022learning,goswami2022deep,zhu2023reliable,subel2023explaining}.

The above methods have successfully demonstrated the capability of neural networks in approximating solution operators. However, in these methods, one model is limited to approximating a single operator, and thus needs to be retrained when the operator changes, which could be due to even a minor change in the PDE.

While the cost of retraining can be reduced by fine-tuning a pretrained model in Act 1~\cite{goswami2020transfer,chen2021transfer,haghighat2021physics,mattheakis2021unsupervised,chakraborty2021transfer,desai2022oneshot,gao2022svd,guo2022analysis,chakraborty2022domain,xu2023transfer} and Act 2~\cite{li2021physics,goswami2022deep,wang2022mosaic,zhu2023reliable,subel2023explaining,xu2023transferdeeponet,lyu2023multi,subramanian2023towards}, we remark that the model must be fine-tuned individually for each distinct function or operator, and such fine-tuning may not be adequate when there are substantial changes to the target function/operator. These drawbacks significantly limit the applicability of the model.

\begin{figure}[htbp]
    \centering
    \includegraphics[width=0.6\linewidth]{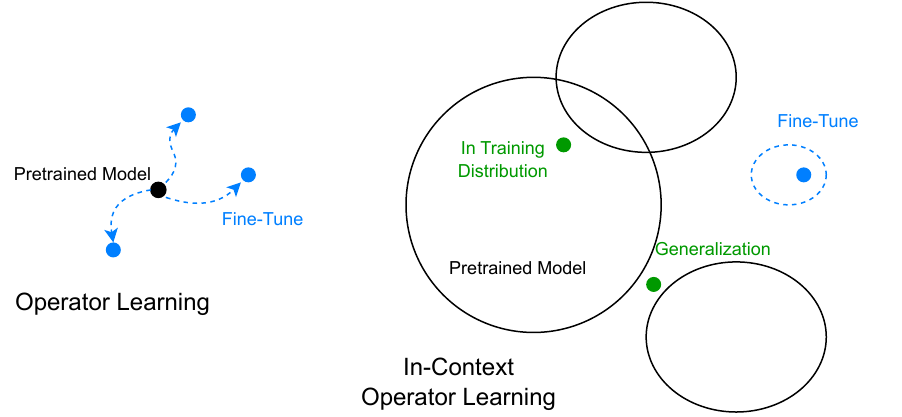}
    \caption{
    Operator Learning v.s. In-Context Operator Learning. For classic operator learning, one model is limited to approximate a single operator, with the need of fine-tuning when the operator changes to a ``close'' new one. For in-context operator learning, a single model can approximate a wide range of operators. It can serve as a ``foundation model'' that could be directly applied without fine-tuning for different PDE-related tasks in the training distribution, or even beyond due to generalization in the operator space. It can also be fine-tuned to strengthen its expertise in particular operator domains, if necessary.
    }
    \label{fig:OLvsICOL}
\end{figure}

In-context operator learning~\cite{yang2023context} can be viewed as the initial attempt of Act 3, where a single model can manage multiple operators. Other examples in Act 3 include ICON-LM~\cite{yang2023prompting} with text and data prompts, PROSE~\cite{liu2023prose} which can generate both the symbolic expression of the governing system and the operator network for multiple distinct ODEs, \cite{liu2023does} for multiple ODEs with in-context operator learning, and Multiple Physics Pretraining (MPP)~\cite{mccabe2023multiple} where a single model learns different physics from different history records.

In-context operator learning approach draws inspiration from in-context learning for Large Language Models (LLMs)~\cite{radford2019language,brown2020language}, where the model performs a task specified by the prompted ``context'', including task descriptions and a few related examples. Recently, the in-context learning paradigm has been applied to other domains, including function regression~\cite{garg2022can}, large vision models~\cite{bai2023sequential}, language-vision models~\cite{liu2023visual}, robotics/embodied models~\cite{driess2023palm}, etc. We refer the readers to
the survey~\cite{dong2022survey} for more details on recent advances in this topic.

For in-context operator learning, the corresponding model ``in-context operator network (ICON)'' acts as an ``operator learner'' rather than being tuned to approximate a specific operator. The trained ICON model is able to infer the operator from the prompted condition-QoI example pairs, and apply the inferred operator to new question conditions for corresponding QoI. Such a learning process happens during the forward pass, without the need for weight adjustments. With different prompted examples, a single ICON model approximates different operators, and thus can tackle a wide range of scientific learning tasks.

For instance, in~\cite{yang2023context}, a single ICON model adeptly manages 19 distinct equations/types of operators, encompassing forward and inverse ordinary differential equations (ODEs), PDEs, and mean-field control problems, with each type containing infinitely many operators and condition/QoI functions being 1D or 2D. In~\cite{yang2023prompting}, ICON is further evolved to take multi-modal prompts, including data examples and texts that describe the task. The demonstrated efficacy hints at the potential for training a ``foundation model'' under the in-context operator learning framework. Such a model could be applied directly for a wide range of PDE-related tasks or, if necessary, be fine-tuned to strengthen its expertise in particular operator domains. We show the comparison between operator learning and in-context operator learning in Figure~\ref{fig:OLvsICOL}.

Large Language Models have demonstrated impressive generalization capabilities, even beyond human expectations~\cite{openai2023gpt4}. In~\cite{yang2023context}, ICON also demonstrated generalization to operators that are not included in the training distribution. However, the generalization is limited to equations with out-of-distribution parameters, and the generalization to new forms of equations is not observed there. Also, \cite{liu2023does} showed the generalization of in-context operator learning in ODEs.

This paper aims to (1) present a detailed methodology for PDE forward and reversal predictions under the in-context operator learning framework, and (2) explore the generalization capabilities of ICON for PDEs. 

In particular, we focus on conservation laws, a family of PDEs with temporal evolution. Our investigation encompasses both time forward and reverse operators. The forward operator takes the initial state as the condition and predicts the system's future state as the QoI, while the reverse operator is conceptualized by making time-reverse predictions, i.e., inverting the roles of the condition and the QoI.

We are motivated to study conservation laws by the following reasons.
\begin{enumerate}
    \item The conservation laws are foundational in describing a wide array of real-world systems.
    \item The family of conservation laws is rich, making it ideal for testing the generalization capabilities of ICON.
    \item The complexities inherent in conservation laws present interesting challenges, including the discontinuities in the solution function, as well as the non-uniqueness of the time-reverse solution.
\end{enumerate}

In this paper, we clearly show that
\begin{center}
  \textit{ICON can generalize well to some PDEs with new forms, without any fine-tuning.}
\end{center}
In particular, we show that an ICON model trained on conservation laws with cubic flux functions can generalize well to some other flux functions of more general forms, without fine-tuning. We believe that this is a significant step towards the goal of training a foundation model for PDE-related tasks under the in-context operator learning framework.

\section{Method}

\subsection{In-Context Operator Network (ICON)}

In this paper, we employ the ICON-LM model introduced in~\cite{yang2023prompting}, which is an improved variant of the ICON model introduced in~\cite{yang2023context}, with ``LM'' stands for ``language model''. Compared with the vanilla ICON model, ICON-LM model exhibits improved training efficiency, achieving better accuracy with about half of the parameters and less training time. As a multi-modal learning model, ICON-LM can also integrate textual human knowledge alongside data examples. However, in this study, we focus on its ability to learn operators solely from data, and for simplicity, we refer to it as ``ICON''.

Denote the neural network as $\T_{\theta}$ with parameters $\theta$. ICON takes a sequence of condition-QoI pairs as input, and predicts each QoI through one forward pass, i.e.
\begin{equation}\label{eqn:forward}
    \{\text{prediction of } \text{QOI}_i \}_{i=2}^I= \T_{\theta}[\{\langle\text{COND}_i, \text{QOI}_i\rangle\}_{i=1}^I].
\end{equation}
Here $\text{COND}_i$ denotes $i$-th condition, and $\text{QOI}_i$ denotes $i$-th QoI. Note that $\{\langle\text{COND}_i, \text{QOI}_i\rangle\}_{i=1}^I$ should be associated with a shared operator, and the operator could be different across different sequences. With a transformer-based architecture and a special attention mask, the prediction of $J+1$-th QoI relies only on the previous condition-QoI pairs (from the first pair to the $J$-th pair), as well as $J+1$-th condition itself. \footnote{Technically, the input sequence consists of condition-QoI-query tuples. These queries are just the inputs of the query functions, indicating where to evaluate the predicted QoI functions. They are conceptually auxiliary, less important than conditions and QoIs, and could be dropped in future variants of ICON. We thus omit them in our notations for simplicity.} This can be written as:
\begin{equation}
    \text{prediction of QOI}_{J+1} = \T_{\theta}[\text{COND}_{J+1};\{\langle\text{COND}_i, \text{QOI}_i\rangle\}_{i=1}^{J}], J = 1,2,...I-1.
\end{equation}
We don't make predictions for $\text{QOI}_1$, since it makes no sense to predict it just with $\text{COND}_1$ without any condition-QoI pairs as examples to indicate the operator.

This approach, termed ``next function prediction'', mirrors the ``next token prediction'' approach in language models. The model is trained by comparing the predicted QoIs and the ground truth QoIs in the sequence, e.g., using the mean squared error as the loss function.

After training, through one forward pass, the ICON model can ingest examples of condition-QoI pairs, learn the corresponding operator, apply it to the question condition, and predict the corresponding QoI. We only need to view the question condition as the last one in the sequence. This relationship can be formulated as:
\begin{equation}
    \text{prediction of QOI}_{\text{question}} = \T_{\theta}[\text{COND}_{\text{question}};\{\langle\text{COND}_i, \text{QOI}_i\rangle\}_{i=1}^J],
\end{equation}
where $J$ is the number of examples used in the prompt for in-context operator learning. A single ICON model can handle different $J$, ranging from one to the maximum capacity $I-1$. For more details of the ICON model, readers are directed to~\cite{yang2023prompting}.

\begin{figure}[htbp]
    \centering
    \includegraphics[width=0.9\linewidth]{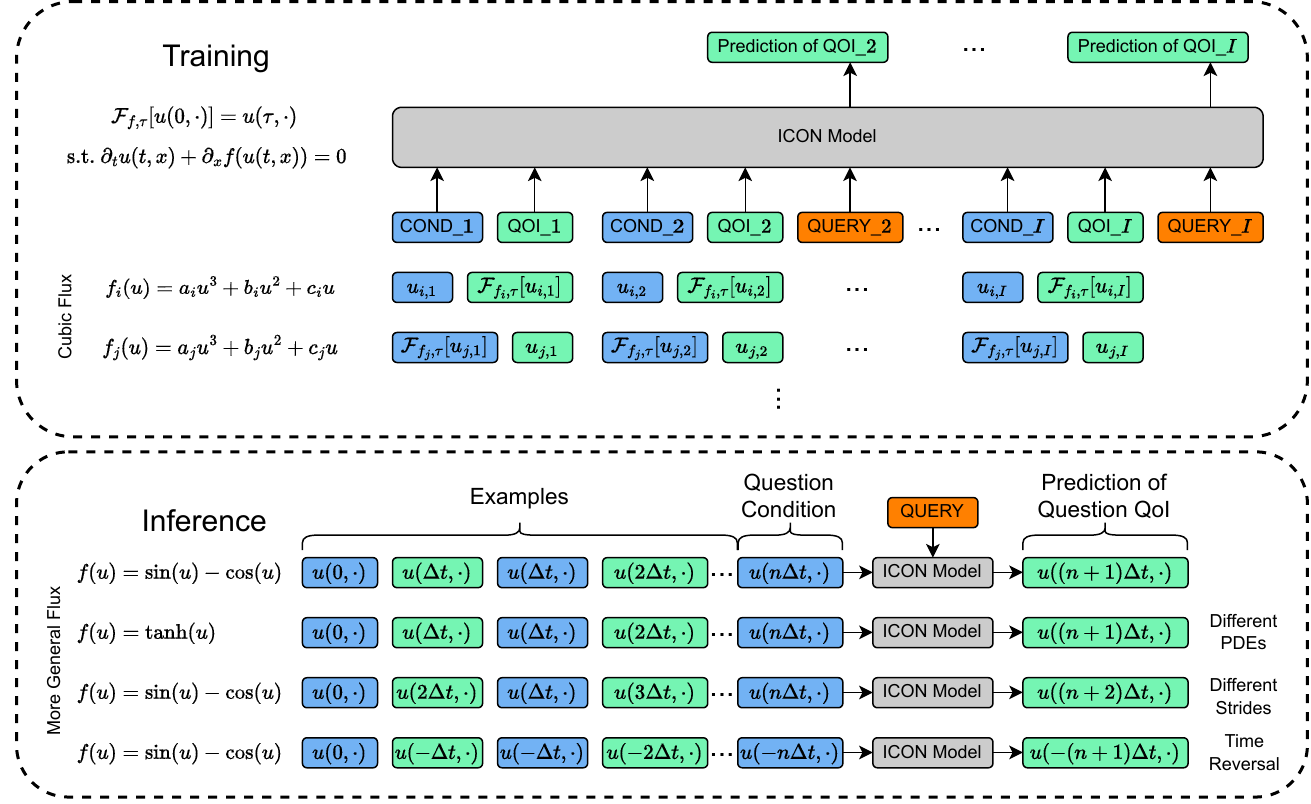}
    \caption{Illuatration of training and inference of ICON for PDEs, using conservation laws as examples.
    }
    \label{fig:illustration}
\end{figure}

\subsection{Forward and Reverse Operators}\label{sec:method_operator}

For PDEs with temporal evolution, the forward operator takes the initial state as the condition and the state at time $\tau$ later as the QoI, while the condition and QoI are swapped in reverse operators. Earlier in~\cite{yang2023context} ICON already demonstrated the capability of in-context operator learning for forward operators in control problems. In this paper, we focus on 1D scalar nonlinear conservation laws that take the following form:
\begin{equation}\label{eqn:conservation_law}
    \partial_t u(t,x)+ \partial_x f(u(t,x)) = 0, x\in[0,1],
\end{equation}
with periodic boundary condition, where $u$ is the solution state, $f$ is the flux. Here we fix the geometry and boundary conditions, so that the forward and reverse operators are defined by the flux $f$ and temporal intervals (or ``stride'' for simplicity) $\tau$. In particular, the forward operator $\mathcal{F}_{f,\tau}$ is defined as
\begin{equation}
\begin{aligned}
   &\mathcal{F}_{f,\tau}[u(0,\cdot)] = u(\tau,\cdot) \\
   \text{s.t.}\quad &\partial_t u(t,x)+ \partial_x f(u(t,x)) = 0
\end{aligned}
\end{equation}
The reverse operator is defined as
\begin{equation}
\begin{aligned}
   \mathcal{R}_{f,\tau}[u] = \{v| \mathcal{F}_{f,\tau}[v] = u\}
\end{aligned}
\end{equation}
Note that the reverse solution is not unique, thus we use the set to include all feasible solutions. 

Each condition or QoI function $u(t,\cdot)$ is represented by a set of key-value pairs, each pair representing a token in the input sequence of the ICON model. In this paper, the key is the spatial coordinate $x$, and the value is the average value of the solution $u$ over the interval $[x, x+\Delta x]$ with $\Delta x = 0.01$, for $x = 0,0.01,...,0.99$, i.e. 100 key-value pairs or tokens for each function.

\subsection{Training}\label{sec:train}

To build the training data, we can simulate different PDEs with different initial conditions and obtain the condition-QoI pairs from the simulation records. Note that one record can be used to generate multiple condition-QoI pairs. 

In particular, in this paper, the training PDEs are 1D scalar conservation laws with cubic flux functions, i.e.,
\begin{equation}\label{eq:burgers}
    \partial_t u + \partial_x (au^3 + bu^2 + cu) = 0,
\end{equation}
with $a, b, c$ uniformly sampled from $[-1, 1]$. We fix the forward and reverse stride $\tau = 0.1$ during training, i.e., the training forward and reverse operators are $\mathcal{F}_{au^3 + bu^2 + cu, 0.1}$ and $\mathcal{R}_{au^3 + bu^2 + cu, 0.1}$ respectively,
with $a, b, c$ uniformly sampled from $[-1, 1]$. The training process is illustrated in Figure~\ref{fig:illustration}.

Recall that the ICON model takes a sequence of condition-QoI pairs as input, and predicts each QoI based on the previous condition-QoI pairs as well as the current condition. It is straightforward to train the model for forward operators: the training loss is defined as the mean squared error between the predicted QoI and the ground truth QoI, i.e. the $L_2$ loss:

\begin{equation}\label{eq:loss_forward}
    \begin{aligned}
        L_\text{Forward}(\theta) =\frac{1}{I-1} \sum_{i=2}^{I}
        \Vert \text{prediction of QOI}_i - \text{QOI}_i \Vert^2  \\
    \end{aligned}
\end{equation}

The reverse operators are more subtle since the solution is not unique. In this paper, we discuss two options:
\begin{enumerate}
    \item \textbf{$L_2$ Loss}: The reverse $L_2$ loss is the same as in Equation~\ref{eq:loss_forward}.
    The only difference is that the condition and QoI are swapped.
    \item \textbf{Consistency Loss}: If we use the model as a surrogate of the exact forward operator, and apply it to the predicted QoI, e.g. $u(0,\cdot)$, ideally we should recover the condition, e.g. $u(\tau,\cdot)$. The consistency loss can be built based on the recovered condition and the ground truth. More formally,
    \begin{equation}\label{eq:loss_consist}
    \begin{aligned}
        L_\text{Consistency}(\theta) &= \frac{1}{I-1} \sum_{i=2}^{I}
        \Vert \hat{\F_i}[ \text{prediction of QOI}_i ] - \text{COND}_i \Vert^2  \\
        \text{where } \hat{\F_i}[\cdot] &=  \T_{\theta}[\cdot;\{\langle\text{QOI}_j, \text{COND}_j\rangle\}_{j=1,j\neq i}^{I}]
    \end{aligned}
\end{equation}
Here we use $\{\langle\text{QOI}_j, \text{COND}_j\rangle\}_{j=1,j\neq i}^{I}$, i.e., all but $i$-th condition-QoI pairs with the condition and QoI swapped, as examples to indicate the forward operator. The parameters $\theta$ are frozen in $\hat{\F_i}$, i.e., the flow of gradients is blocked, to reduce the computational cost.

\end{enumerate}

Mathematically, the reverse $L_2$ loss aims to find the expectation of multiple feasible solutions, which strictly speaking may not be a feasible solution. The consistency loss is more rigorous but also computationally more expensive. Moreover, the effectiveness of the consistency loss heavily depends on the accuracy of the forward operator surrogate. We compare the two options in Section~\ref{sec:results_ind}.

In the end, we remark that we didn't use automatic differentiation in the loss function to incorporate the PDE information. Instead, ICON learns from data alone, in particular, generated by the third-order Weighted Essentially Non-Oscillatory (WENO) scheme~\cite{liu1994weighted} in this paper. This strategy not only reduces the training costs but also significantly enhances the robustness of the training process, especially when dealing with discontinuities. Indeed, the battle-tested stability and accuracy of the numerical scheme play a critical role in handling the discontinuities in the solutions of conservation laws. The details of data generation are presented in Section~\ref{sec:data}.

\subsection{Inference}
After training, a single ICON model can be applied to various PDEs or condition functions. During such an inference phase, there are no weight updates at all, thus the computational cost is extremely small. The model's adaptability comes from the way we construct the data prompts, including the example condition-QoI pairs and the question condition, similar to designing prompts for a language model.

We emphasize that the PDE is not explicitly fed into the model via parameters or other means. Instead, the model learns the PDE implicitly from the condition-QoI examples. Therefore, the model can naturally be applied to PDEs with new forms, as long as the condition-QoI examples are constructed correspondingly. In this paper, we train the model with cubic flux functions, and test the model with flux functions of more general forms. Also, while the time stride $\tau$ is fixed in training, it can vary in inference, since this is equivalent to scaling the PDE properly, as we will discuss in Section~\ref{sec:stride}. The inference phase is illustrated in Figure~\ref{fig:illustration}.

Based on the source of the condition-QoI examples, there are three cases:

\begin{enumerate}
    \item \textbf{Self Reference}: In this case, for data we have a sequence 
    \begin{equation}\label{eqn:seq}
    ( u(0), u(\Delta t), u(2\Delta t), \cdots, u(n\Delta t)  ) =: (u(i\Delta t))_{i=0}^n,
    \end{equation}
    where $u(t)$ is the solution of the PDE at time $t$. We call the sequence a ``record'', and each $u(t)$ a ``frame''. We will construct condition-QoI examples from the given record, and predict future frames, i.e. $u((n+1)\Delta t), u((n+2)\Delta t), \cdots$, or previous frames, i.e. $u(-\Delta t), u(-2\Delta t), \cdots$.
    \item \textbf{Single Reference Record}: In this case, we have a reference record and make predictions for a new initial/terminal condition.
    \item \textbf{Multiple Reference Records}: In this case, we have multiple reference records governed by the same PDE, and make forward/reverse predictions for a given initial/terminal condition.
\end{enumerate}

``Self Reference'' can be viewed as a special case of ``Single Reference Record'', but we list it separately since it is a very common case in practice. For all these cases, we can construct a condition-QoI example by randomly sampling a frame from a record as the condition, and a later/previous frame as the QoI from the same record with a certain stride. 

After the introduction of ICON, other researchers also looked into training one model for multi-physics prediction, by specifying physics with history records~\cite{mccabe2023multiple}. In particular, in the inference stage, the model input is the sequence $(u(i\Delta t))_{i=0}^n$
which implicitly encodes the physics, and the output is the prediction of $u((n+1)\Delta t)$. The model is used in an auto-regressive manner to make predictions for $u((n+2)\Delta t), u((n+3)\Delta t), \cdots$. We refer to this approach as ``video prediction'' since it draws a parallel between forecasting PDE and predicting video frames.

While this approach is capable of addressing multiple PDEs, it overlooks an important property: common PDEs are time-homogeneous Markovian processes, i.e. the future state only depends on the current state, and such dependence is invariant in time\footnote{Some terms in the PDE can be time-dependent, but usually these terms are the system states or controls, and the PDE that governs the system states or controls is time-homogeneous Markovian.}. This property is not only a key feature of most PDEs, but also forms the foundation of both forward and reverse operator definitions, as well as classic numerical PDE schemes.

By taking advantage of the time-homogeneous Markovian property of PDEs, and constructing condition-QoI examples in a very flexible way, ICON is more powerful. 

If we construct the condition-QoI examples as 
\begin{equation}
  \langle u(0), u(\Delta t)\rangle, \langle u(\Delta t), u(2\Delta t)\rangle, \cdots, \langle u((n-1)\Delta t), u(n\Delta t)\rangle,
\end{equation}
and set $u(n\Delta t)$ as the question condition, then the ICON model will recover the video prediction approach and predict $u((n+1)\Delta t)$ . Here we use the notation $\langle u(t_1), u(t_2) \rangle$ to denote a condition-QoI example, where $u(t_1)$ is the condition and $u(t_2)$ is the QoI.

Beyond one-step prediction, ICON can easily make multi-step or large-stride predictions by simply changing the order of frames in the input sequence. For example, suppose $n$ is a multiple of 2 and we want to make predictions for $u((n+n/2)\Delta t)$, we can construct condition-QoI examples as 
\begin{equation}
  \langle u(0), u(n/2\Delta t)\rangle, \langle u(\Delta t), u((n/2+1)\Delta t)\rangle, \cdots, \langle u(n/2\Delta t), u(n\Delta t)\rangle,
\end{equation}
and set $u(n\Delta t)$ as the question condition, then the model will predict $u((n+n/2)\Delta t)$ with one forward pass.

To make reverse predictions, given the sequence $(u(-i\Delta t))_{i=0}^n$, we can construct condition-QoI examples as 
\begin{equation}
  \langle u(0), u(-\Delta t)\rangle, \langle u(-\Delta t), u(-2\Delta t)\rangle, \cdots, \langle u(-(n-1)\Delta t), u(-n\Delta t)\rangle,
\end{equation}
and set $u(-n\Delta t)$ as the question condition, so the ICON model will learn the reverse operator and predict $u(-(n+1)\Delta t)$. Similarly for multi-step reverse predictions.

In the end, we remark that ICON is adaptive to non-Markovian processes as well: we just need to set the condition as multiple frames instead of one.

\subsection{Recursive Predictions}\label{sec:recursive}

To make predictions for a long time horizon, we can recursively call the ICON model with the predicted QoIs as the new question conditions. Since a single ICON model can make predictions with different strides, there are multiple schemes to recursively make predictions. We present one example here, which is used throughout the paper.

\begin{figure}[htbp]
    \centering
    \includegraphics[width=0.5\linewidth]{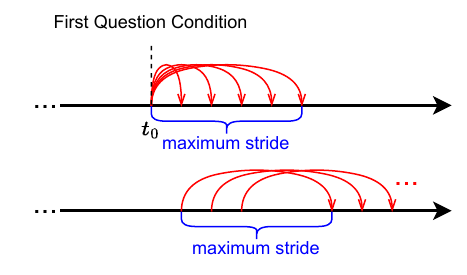}
    \caption{Illustration of the scheme for recursive predictions.}
    \label{fig:recursive_scheme}
\end{figure}

Given the record(s) for building condition-QoI examples, as well as $u_{t_0}$ as the first question condition, we make recursive forward predictions with the following steps:
\begin{enumerate}
    \item Prescribe the maximum stride $S$.
    \item Using $u(t_0)$ as the question condition, make predictions of $u(t_0 + s)$ with the stride $s$, for $s = \Delta t,2\Delta t,...,S$.
    \item Using $u(t-S)$ as the question condition, make predictions of $u(t)$ with the maximum stride $S$, for $t = t_0+S +\Delta t, t_0+S + 2\Delta t, \cdots$.
\end{enumerate}

An illustration of the recursive prediction scheme is shown in Figure~\ref{fig:recursive_scheme}. It's clear that we can make predictions for $t_0+n\Delta t$ for $n\in \mathbb{N}$ in this way. The recursive reverse predictions can be made similarly.

\subsection{Transform Operators and Functions to ICON's Capability Scope}

ICON can handle a range of operators and condition functions through in-context operator learning. When faced with unfamiliar operators and/or condition functions, it's sometimes possible to transform them into ones that ICON can effectively process. Such techniques broaden the range of problems that ICON can address and strengthen its role as a foundation model. As an analogy, the concept of ``chain of thought''~\cite{wei2022chain} has been introduced to decompose complex reasoning tasks into smaller, more manageable segments that can be tackled by a Large Language Model.

Here, we demonstrate two simple examples of such techniques: change of variables and varying strides. 

\subsubsection{Change of Variables}
When applying the change of variables, both the equation and condition function are transformed. Consider the conservation law $\partial_t u + \partial_x f(u) = 0$, take a simple affine transformation as example, i.e., $u = \alpha v + \beta$ with $\alpha>0$, then the equation becomes
\begin{equation}
\begin{aligned}
    \partial_t v +  \partial_x f(\alpha v + \beta)/\alpha = 0,
\end{aligned}
\end{equation}
In other words, the condition function is transformed to $v = (u-\beta)/\alpha$, and the flux function is transformed to $f(\alpha v + \beta)/\alpha$. For condition functions and flux functions that are beyond the training scale, we can apply the affine transformation with $\alpha > 1$, so that both can be scaled down to the training scale.

Here we remark that the prediction error is not proportional to the scale of the condition function or flux function. If $\alpha$ is too large, although the condition function and flux function could be scaled down to the training scale, the prediction error will be amplified when transforming back to the original variables. The detailed numerical results are presented in Section~\ref{sec:results_variable}.

The training PDEs are limited to conservation laws with cubic flux functions in this paper, and we thus only considered linear transform. If the training distribution covers more general cases, more sophisticated transformations can also be effective.

\subsubsection{Varying Strides}\label{sec:stride}
For conservation laws of form~\ref{eqn:conservation_law}, we can see that 
\begin{equation}
  \mathcal{F}_{kf,\tau} = \mathcal{F}_{f,k\tau},  \quad \mathcal{R}_{kf,\tau} = \mathcal{R}_{f,k\tau},
\end{equation}
where $k$ is a positive constant. In other words, if we apply a smaller/larger stride, it is equivalent to solving a PDE with a smaller/larger flux and the original stride. This technique can be used to adjust the scale of flux functions to the desired range. 

Note that with a smaller stride, it takes more steps to reach the same time, i.e., more calls to the ICON model. This could lead to a larger error accumulation. Therefore we need to make a tradeoff between the generalization ability of ICON and the error accumulation. The detailed numerical results are presented in Section~\ref{sec:results_time}.

\section{Data Preparation}\label{sec:data}

The training data are generated by simulating conservation laws with cubic flux functions:
\begin{equation}
    \partial_t u + \partial_x (au^3 + bu^2 + cu) = 0, \quad x \in [0,1],
\end{equation}
with periodic boundary conditions. The simulation is conducted with the third-order WENO scheme~\cite{liu1994weighted}, a battle-tested numerical scheme for conservation laws.

For the training phase, we employ a set of 1000 tuples of $(a,b,c)$, each randomly sampled from the hypercube $ [-1, 1]^3 $. The following protocol is adopted for each tuple of operator parameters $(a, b, c)$:

\begin{enumerate}
    \item \textbf{Initial Conditions:} Sample $ N = 100 $ periodic functions as initial conditions. These functions are defined on a grid with spacing $ \Delta x = 0.01 $. In practice, we sample these functions from a periodic Gaussian random field with zero mean and covariance kernel 
    \begin{equation}
    \begin{aligned}
    k(x,x') & =  \sigma^2 \exp\left(-\frac{1-\cos(2\pi (x-x'))}{l^2}\right), 
    \end{aligned}
    \end{equation}
  where $\sigma = 1, l = 1$. The initial functions with value beyond $[-3,3]$ are dropped for the sake of the CFL condition during data generation.
    \item \textbf{Numerical Solution:} Employ the third-order WENO scheme and the fourth-order Runge-Kutta method to solve the conservation law. The system evolves from $ t=0 $ to $ t=0.5 $ using a time step $ \Delta t = 0.0005 $, resulting in 1001 steps in total, taking account of both the initial and final states. The solution we get is the weak solution which satisfies the jump condition for the discontinuities, and also satisfies the entropy condition. 
    \item \textbf{Data Collection:} Consider each of the first 801 time steps corresponding to the initial \( 0.4 \) time units as an individual initial condition. Each of these has a corresponding function that occurs $ 0.1 $ time units later. Such pairs of functions can be collected as the conditions and QoIs for the forward and reverse operators.
\end{enumerate}

Given $ N $ initial functions, this procedure yields a total of $ 801N = 80100$ conditions-QoIs pairs for each operator.  To optimize storage efficiency, we do not exhaustively utilize all generated condition-QoI pairs. Instead, for each operator we randomly down-sample and store $100N = 10000 $ pairs for training.

\section{Experimental Results} \label{sec:results}

The ICON model employs a transformer architecture, in this paper configured as in Table~\ref{tab:transformer-config}. For optimization, the AdamW optimizer is used in conjunction with a warmup-cosine-decay schedule, following the parameters set in Table~\ref{tab:optimizer-config}.

The input and output layers are linear layers. In this paper, the input sequence during training consists of six condition-QoI pairs. Consequently, the inference phase is limited to a maximum of five examples, plus a single question condition. The condition/QoI functions and the five additional queries related to QoIs (excluding the first one) comprise 100 tokens each, as is explained in Section~\ref{sec:method_operator}. This results in a total of 1700 tokens in the input sequence during training. 

\begin{table}[ht]
\centering
\caption{Transformer Configuration}
\begin{tabular}{|c|c|}
\hline
{Layers} & 6 \\
\hline
{Heads in Multi-Head Attention} & 8 \\
\hline
{Input/Output Dimension of Each Layer} & 256 \\
\hline
{Dimension of Query/Key/Value in Attention Function} & 256 \\
\hline
{Hidden Dimension of Feedforward Networks} & 1024 \\
\hline
\end{tabular}
\label{tab:transformer-config}
\end{table}

\begin{table}[ht]
\centering
\caption{Configuration of Optimizer and Learning Rate Schedule}
\begin{tabular}{|c|c|}
\hline
{Initial Learning Rate} & 0.0 \\
\hline
{Peak Learning Rate} & 1e-4 \\
\hline
{End Learning Rate} & 0.0 \\
\hline
{Training Steps} & $10^6$\\
\hline
{Warmup Steps} & First 10\% of Total Steps\\
\hline
{Cosine Annealing Steps} & Remaining Steps\\
\hline
{Global Norm Clip} & 1.0\\
\hline
{Adam $\beta_1$} & 0.9\\
\hline
{Adam $\beta_2$} & 0.999\\
\hline
{Adam Weight Decay} & 1e-4\\
\hline
\end{tabular}
\label{tab:optimizer-config}
\end{table}

\subsection{Metrics}\label{sec:results_metric}
Denote the trained neural network as $\T_{\theta}$ as in Equation~\ref{eqn:forward}, then given some examples $\{\langle u_i; \F_{f,\tau}[u_i] \rangle \}_i^I$, we can approximate the forward operator $\F_{f,\tau}$ with $\T_{\theta}[\cdot;\{\langle u_i, \F_{f,\tau}[u_i] \rangle\}_i^I]$, and the reverse operator $\mathcal{R}_{f,\tau}$ with $\T_{\theta}[\cdot;\{\langle \F_{f,\tau}[u_i], u_i\rangle \}_i^I]$.

We denote 
\begin{equation}
\begin{aligned}
    \hat{\F}_{f,\tau}[u] &:= \T_{\theta}[u;\{\langle u_i, \F_{f,\tau}[u_i] \rangle\}_i^I], \\
    \hat{\mathcal{R}}_{f,\tau}[u] &:= \T_{\theta}[u;\{\langle \F_{f,\tau}[u_i], u_i\rangle \}_i^I].
\end{aligned}
\end{equation}
The condition-QoI examples $\{\langle u_i, \F_{f,\tau}[u_i] \rangle\}_i^I$ and $\{\langle \F_{f,\tau}[u_i], u_i\rangle \}_i^I$ are dropped for simplicity. We will specify them as needed in the following sections.

To quantify the performance of these learned operators, we define two metrics: the forward error and the reverse error. The forward error is straightforward, which is the average $L_1$ distance between the predicted QoI and the ground truth QoI:
\begin{equation}\label{eq:ferror}
    \text{Forward Error} := \| \hat{\F}_{f,\tau}[u] - \F_{f,\tau}[u] \|_1,
\end{equation}
The reverse error is more subtle since the solution is not unique. We thus apply the exact forward operator~\footnote{Here we use forward simulation to apply the ``exact'' forward operator. Strictly speaking, such an operator also has numerical errors and thus is not ``exact''. However, the errors are negligible compared with neural network prediction errors, and we thus ignore them.} to the predicted QoI, and compare the result with the condition function:
\begin{equation}\label{eq:berror}
    \text{Reverse Error} := \| \F_{f,\tau}[\hat{\mathcal{R}}_{f,\tau}[u]] - u \|_1.
\end{equation}

For recursive forward predictions, the error is also defined as the average $L_1$ distance between the predictions and the ground truth. For recursive reverse predictions, we apply the exact forward operator to the predictions until $t_0$, the time for the first question condition,
and calculate the $L_1$ distance between the reconstructed first question condition and the ground truth.

\subsection{In-Distribution Operators}\label{sec:results_ind}

In this section, we evaluate the performance of ICON on in-distribution operators, specifically focusing on the forward operators $\F_{au^3 + bu^2 + cu,0.1}$ and reverse operators $\mathcal{R}_{au^3 + bu^2 + cu,0.1}$, with the number of examples ranging from 1 to 5. Here $(a,b,c)$ are on the $11\times 11 \times 11$ uniform grid within $[-1,1]^3$. For each operator, the condition-QoI pairs are generated in the same way as for training.

We also compared the $L_2$ loss and consistency loss for training reverse operators. In particular, we consider three training configurations:
\begin{enumerate}
    \item $L_2$ loss with batch size 4 for forward operators and consistency loss with batch size 4 for reverse operators. The training takes about 45.5 hours on dual NVIDIA RTX 4090 GPUs.
    \item $L_2$ loss with expected batch size 8 for forward operators and another 8 for reverse operators. \footnote{When using $L_2$ loss for both forward and reverse operators, practically we train the model with data randomly sampled from forward and reverse operators. The exact split in each iteration may be different, but the expectation is half-half. When using consistency loss, in each iteration, we use the same 4 sequences to build the forward $L_2$ loss and the consistency loss.} The training takes about 40 hours on dual NVIDIA RTX 4090 GPUs.
    \item $L_2$ loss with expected batch size 4 for forward operators and another 4 for reverse operators. The training takes about 37 hours on a single NVIDIA RTX 4090 GPU.
\end{enumerate}

\begin{figure}[htbp]
    \centering
    \begin{subfigure}{0.45\textwidth}
    \centering
    \includegraphics[width=\linewidth]{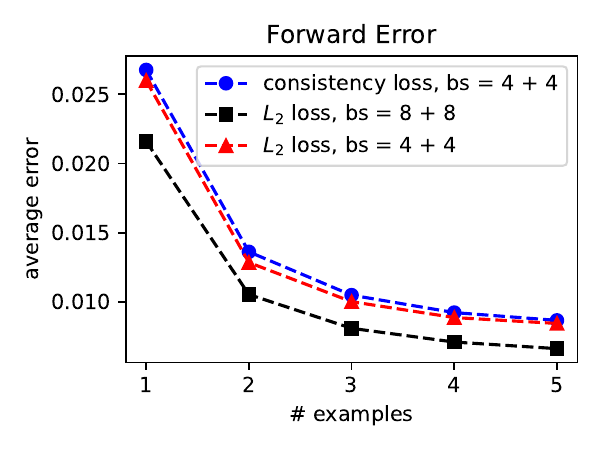}
    \caption{}
    \label{fig:results_decay_1}
    \end{subfigure}
    \begin{subfigure}{0.45\textwidth}
    \centering
    \includegraphics[width=\linewidth]{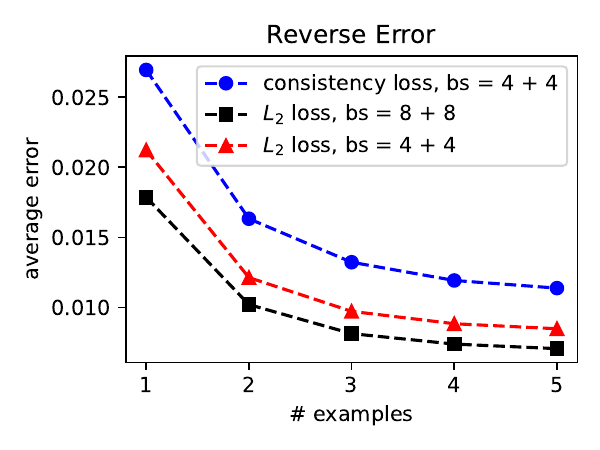}
    \caption{}
    \label{fig:results_decay_2}
    \end{subfigure}
    \caption{Averaged error v.s. the number of examples used for in-context operator learning. (a) Forward error. (b) Reverse error.}
    \label{fig:results_decay}
\end{figure}

In Figure~\ref{fig:results_decay}, we show the forward error and reverse error averaged over the operators on the grid and 100 instances of in-context operator learning for each operator. For all three configurations, it's clear that both errors decay as the number of in-context examples increases. The consistency loss performs the worst in our experiments, although with the largest computational cost. This may be due to that the model cannot serve as an accurate forward operator surrogate before convergence, and thus the consistency loss is not effective. A more sophisticated training strategy can be developed to improve the performance of consistency loss, e.g., gradually increasing the weight of consistency loss as the model converges. Since this is not the focus of this paper, we leave it as future work.

\begin{figure}[htbp]
    \centering
    \includegraphics[width=0.24\linewidth]{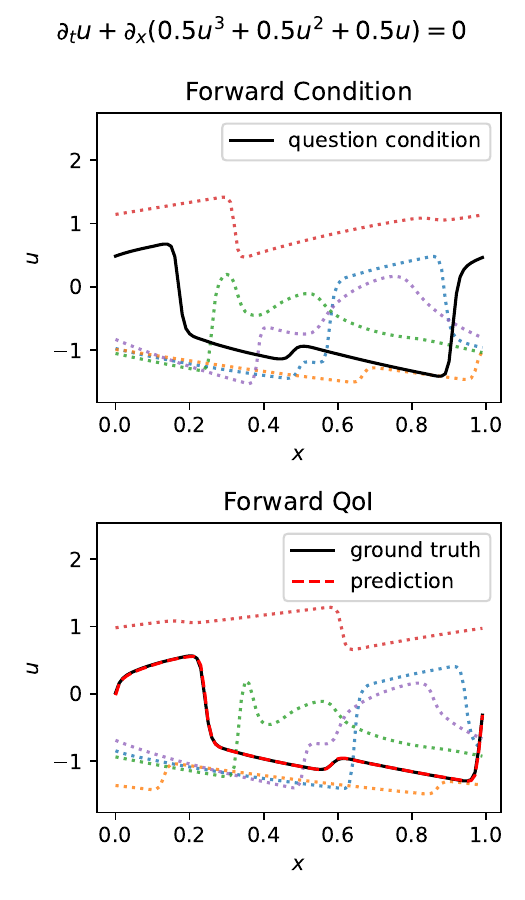}
    \includegraphics[width=0.24\linewidth]{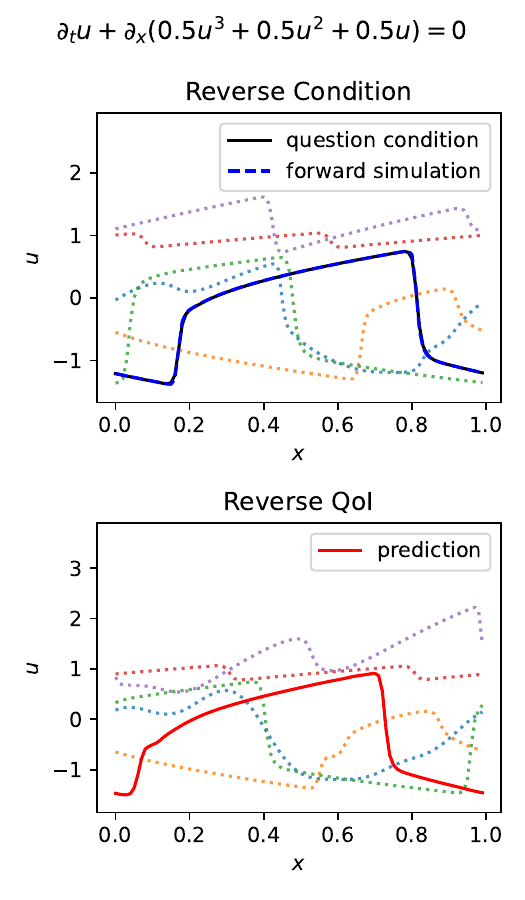}
    \includegraphics[width=0.24\linewidth]{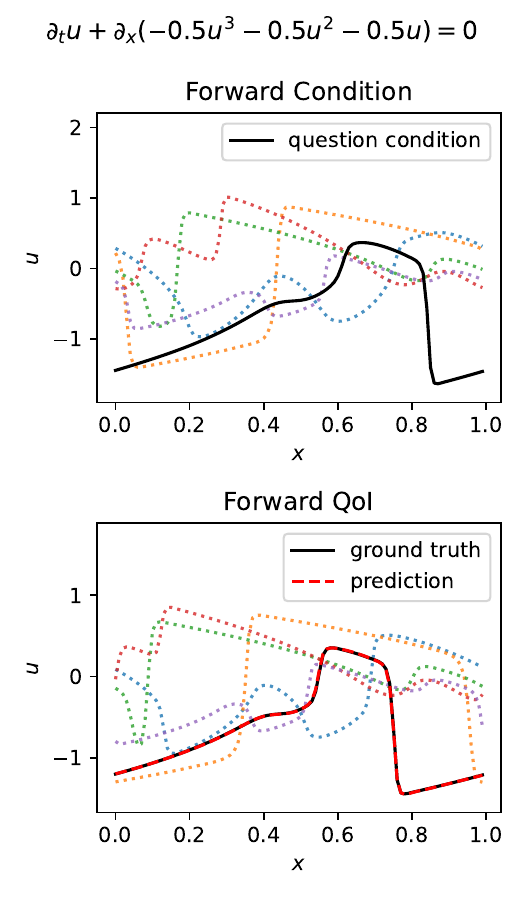}
    \includegraphics[width=0.24\linewidth]{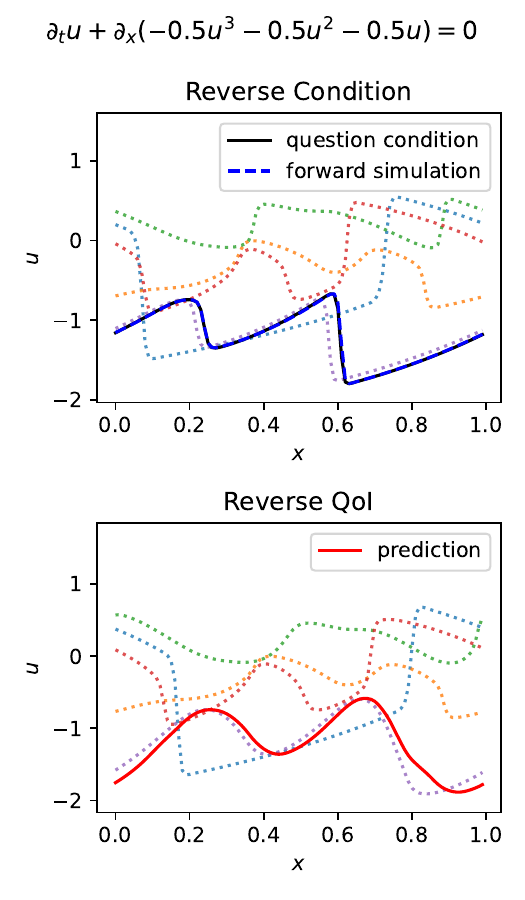}
    \caption{Illuatration of in-context operator learning for $\F_{0.5u^3 + 0.5u^2 + 0.5u,0.1}$, $\mathcal{R}_{0.5u^3 + 0.5u^2 + 0.5u,0.1}$, $\F_{-0.5u^3 -0.5u^2 -0.5u,0.1}$ and $\mathcal{R}_{-0.5u^3 -0.5u^2 -0.5u,0.1}$. For each case, the prompted five condition-QoI examples are shown with dotted color lines. The forward predictions shown with dashed red lines overlap with the ground truth QoIs shown with solid black lines. Since there are no unique ground truth solutions for the reverse operators, we apply the exact forward operators to the predicted QoIs by forward simulation, and show the recovered conditions with dashed blue lines, which overlap with the question conditions shown with solid black lines.}
    \label{fig:results_cubic_profile}
\end{figure}

Since the $L_2$ loss with $8+8$ batch size works the best, from now on, we will analyze the results using the model trained with this configuration. As an illustration, in Figure~\ref{fig:results_cubic_profile} we show some cases of in-context forward and reverse operator learning, corresponding to different equations. The overlapping between the predicted QoI and the ground truth QoI for forward operators, and the overlapping between the recovered condition and the question condition for reverse operators, indicate that the learned operators are accurate.

\subsection{Generalization to New PDEs}\label{sec:generalization}
In this section, we show that the ICON can generalize to new PDEs with more general forms of flux functions. 

We apply ICON recursively to make predictions for a long time horizon. In particular, for forward predictions, we consider that we have 11 frames of data $u(0), u(0.01), \cdots, u(0.1)$, and use the trained ICON to predict $u(0.11), u(0.12) \cdots, u(0.5)$. We follow the recursive scheme introduced in Section~\ref{sec:recursive}, with $t_0 = 0.1$, $\Delta t = 0.01$, and the maximum stride $S = 0.05$. For all predictions, we use five condition-QoI pairs as prompted examples, which are randomly sampled from 
\begin{equation}
    \langle u(0), u(s)\rangle, \langle u(0.01), u(0.01 + s)\rangle, \cdots, \langle u(0.1-s), u(0.1)\rangle,
\end{equation}
when making predictions with a stride of $s$.

\begin{figure}[htbp]
    \centering
    \includegraphics[width=0.95\linewidth]{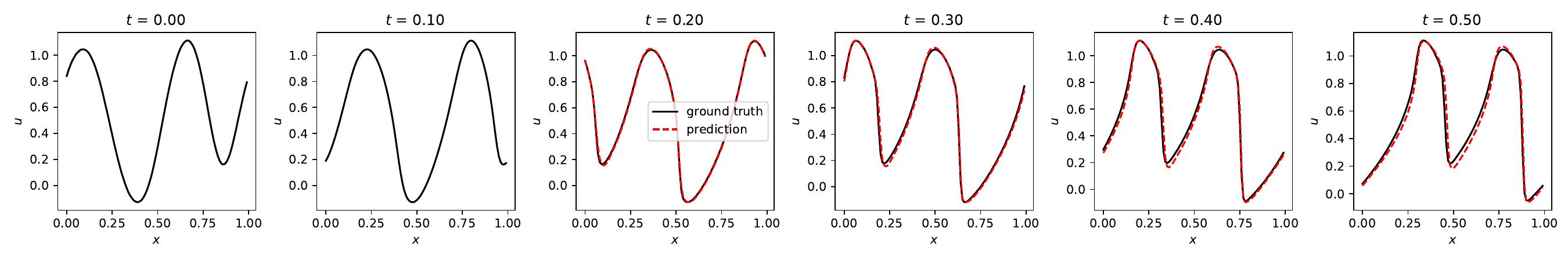}
    \includegraphics[width=0.95\linewidth]{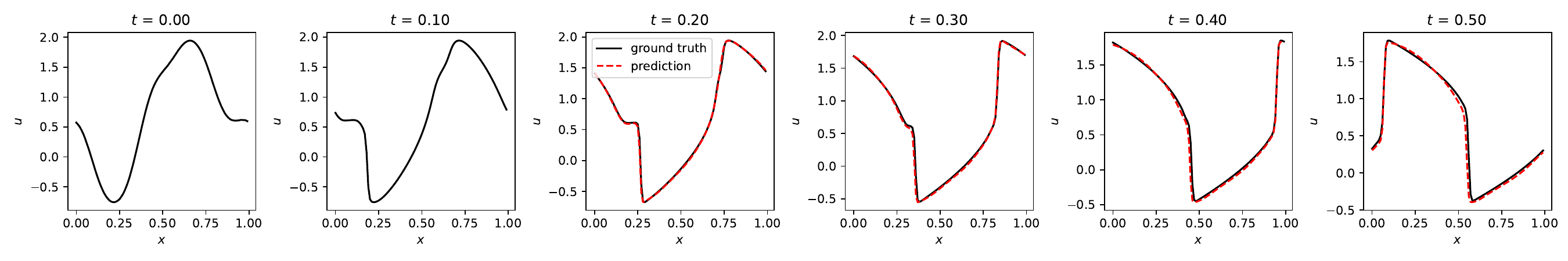}
    \includegraphics[width=0.95\linewidth]{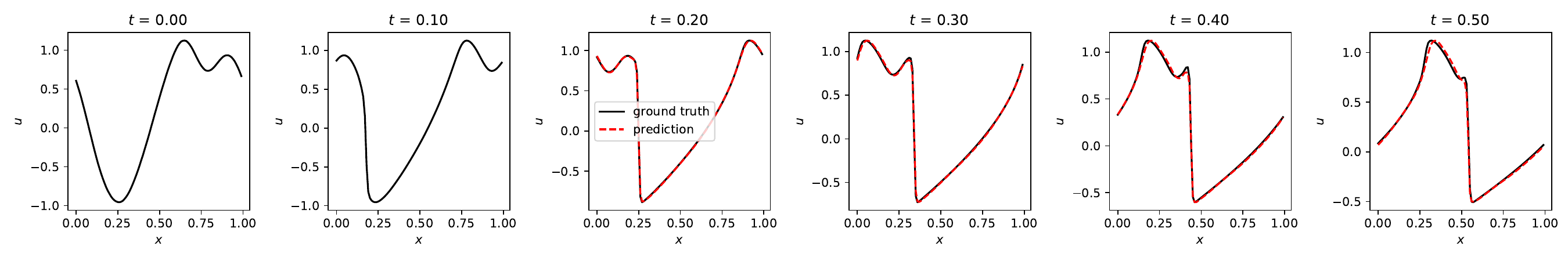}
    \caption{Three examples of forward prediction of $\partial_t u + \partial_x(\sin(u)-\cos(u)) = 0$. The overlapping between the forward predictions and the ground truth data indicates that the forward predictions are accurate.}
    \label{fig:results_new_forward_profile}
\end{figure}

\begin{figure}[htbp]
    \centering
    \includegraphics[width=0.95\linewidth]{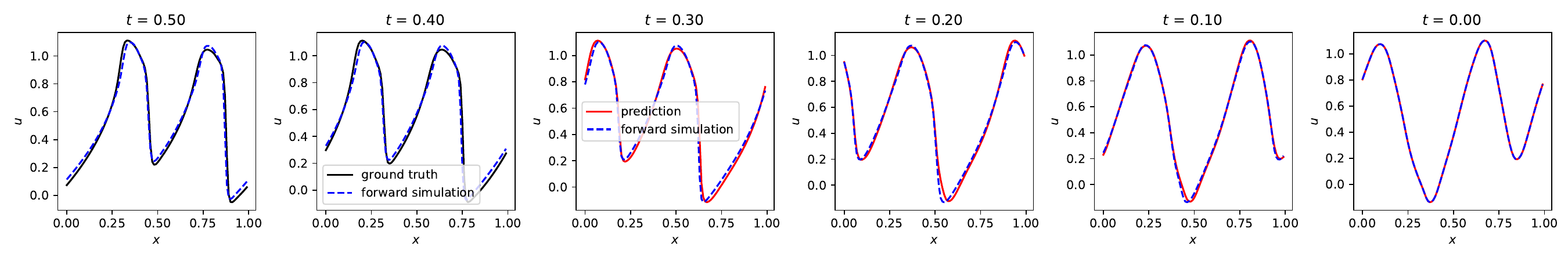}
    \includegraphics[width=0.95\linewidth]{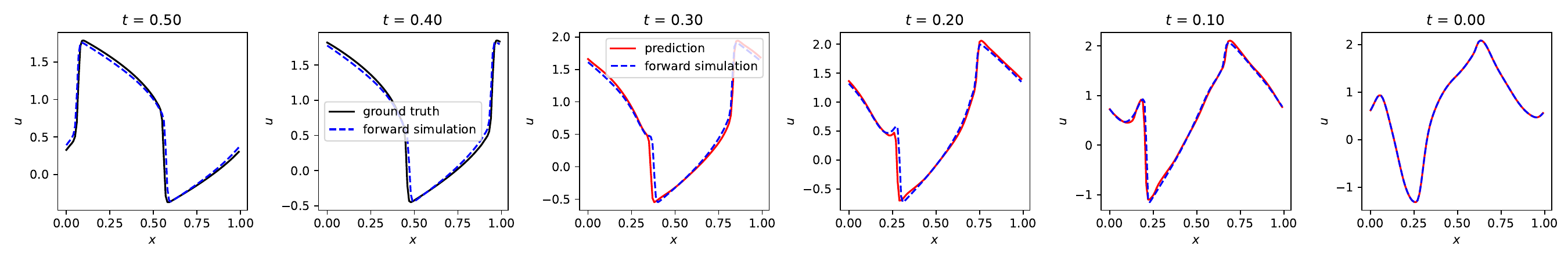}
    \includegraphics[width=0.95\linewidth]{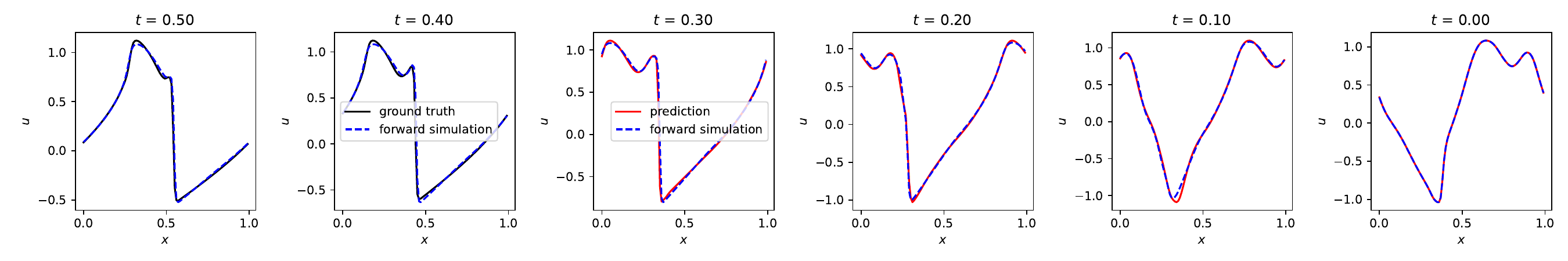}
    \caption{Three examples of reverse prediction of $\partial_t u + \partial_x(\sin(u)-\cos(u)) = 0$. We apply the exact forward simulation from the predicted initial condition at $t = 0$ until $t = 0.5$, shown with blue dashed lines. The overlapping between the forward simulation results and ground truth data indicates that the reverse predictions are accurate.}
    \label{fig:results_new_backward_profile}
\end{figure}

Similarly, for reverse predictions, we consider that we have 11 frames of data $u(0.5), u(0.49), \cdots, u(0.4)$, and use the trained ICON to predict $u(0.39), u(0.38) \cdots, u(0)$, with $t_0 = 0.4$.

In Figure~\ref{fig:results_new_forward_profile} and~\ref{fig:results_new_backward_profile} we showcase some forward and reverse prediction results for $\partial_t u + \partial_x(\sin(u)-\cos(u)) = 0$. We can see the great accuracy, even for the PDE and initial functions that are never seen during training. 

How accurate is the prediction for the new PDE compared with the predictions for PDEs with cubic flux functions? Did the ICON model simply memorize the cubic flux functions and approximate the new flux with the closest cubic function?

To answer these questions, we consider the following two comparisons:
\begin{enumerate}
    \item[] Comparison 1: We compare the errors of predictions for different equations, including $f(u) = \sin(u)-\cos(u)$ and ``similar'' cubic flux functions. The definition of ``similar'' will be introduced later.
    \item[] Comparison 2: We use different equations, including $f(u) = \sin(u)-\cos(u)$ and ``similar'' cubic flux functions, to generate the prompted examples. We make predictions with these examples, and compare the errors between the predictions and the ground truth corresponding to $f = \sin(u)-\cos(u)$. To ensure a fair comparison, the initial condition for generating prompted examples via simulation, as well the first question condition $u(t_0)$ for recursive predictions, are shared across all equations.
\end{enumerate}

The ``similar'' cubic flux functions consist of the following:
\begin{enumerate}
    \item The cubic Taylor polynomial of $f(u) = \sin(u)-\cos(u)$ at $0$, i.e. $f(u) = -1/6 u^3 + 1/2u^2 + u $. The constant term is dropped since it does not affect the solution. Same for the following.
    \item The best cubic fit of $f(u) = \sin(u)-\cos(u)$ within $[-1,1]$ in the $L_2$ sense, i.e., $f(u) = -0.157u^3 + 0.465 u ^2 + 0.998 u$ approximately.
    \item The best cubic fit of $f(u) = \sin(u)-\cos(u)$ within $[-2,2]$ in the $L_2$ sense, i.e., $f(u) = -0.132u^3 + 0.370 u ^2 + 0.971 u$ approximately.
    \item Since the range of $u$ varies for PDE with different initial conditions, we also consider the best cubic fit of $f(u) = \sin(u)-\cos(u)$ within $[u_{\min},u_{\max}]$ in the $L_2$ sense, where $u_{\min}$ and $u_{\max}$ are the minimum and maximum values of $u$ in the initial condition. Due to the maximum principle, the range of $u$ will not exceed $[u_{\min},u_{\max}]$ for all time. We denote this as ``adaptive cubic fit''.
\end{enumerate}

\begin{figure}[htbp]
    \centering
    \begin{subfigure}{0.4\textwidth}
        \centering
        \includegraphics[width=\linewidth]{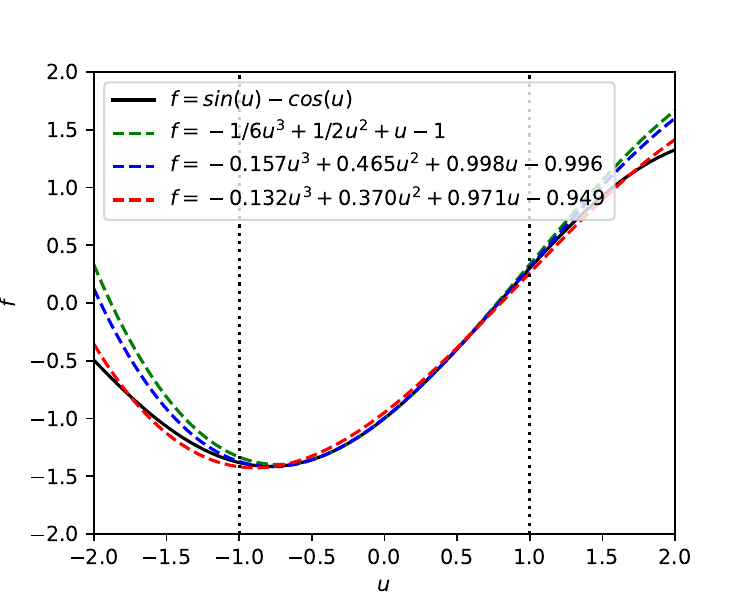}
        \caption{}
        \label{fig:results_new_fit_1}
        \end{subfigure}
        \begin{subfigure}{0.4\textwidth}
        \centering
        \includegraphics[width=\linewidth]{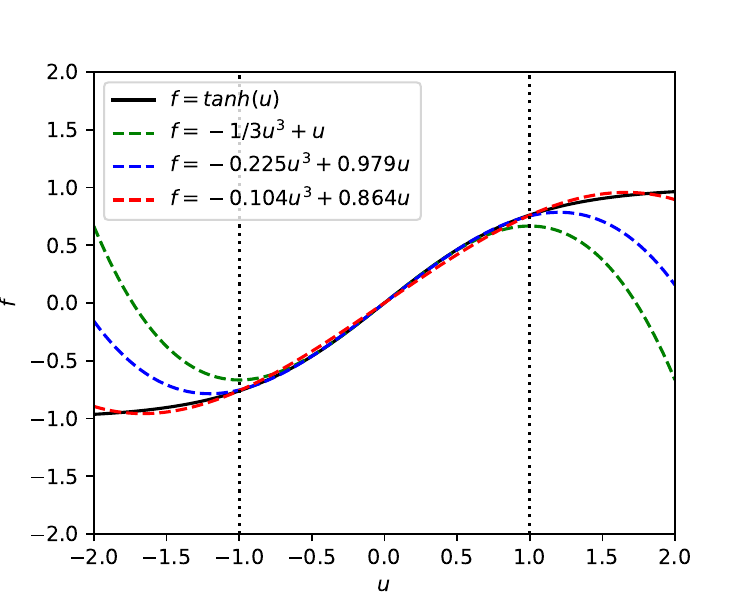}
        \caption{}
        \label{fig:results_new_fit_2}
        \end{subfigure}
    \caption{The tested flux functions (black solid lines) and ``similar'' cubic functions (colored dashed lines), including cubic Taylor polynomials (green), cubic fit in $[-1,1]$ (blue), and cubic fit in $[-2,2]$ (red). (a) $f = \sin(u) - \cos(u)$. (b) $f = \tanh(u)$.}
    \label{fig:results_new_fit}
\end{figure}

The first three ``similar'' cubic functions are illustrated in Figure~\ref{fig:results_new_fit_1}.

\begin{figure}[htbp]
    \centering
    \begin{subfigure}{0.24\textwidth}
    \centering
    \includegraphics[width=\linewidth]{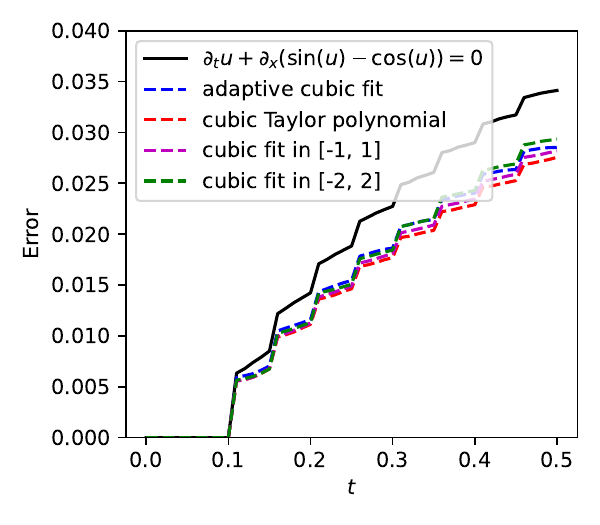}
    \caption{}
    \label{fig:results_new_cubicfit_sin_1}
    \end{subfigure}
    \begin{subfigure}{0.24\textwidth}
    \centering
    \includegraphics[width=\linewidth]{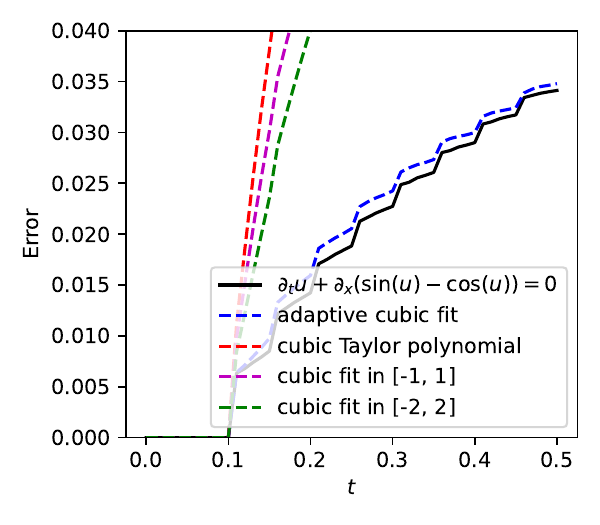}
    \caption{}
    \label{fig:results_new_cubicfit_sin_2}
    \end{subfigure}
    \begin{subfigure}{0.24\textwidth}
    \centering
    \includegraphics[width=\linewidth]{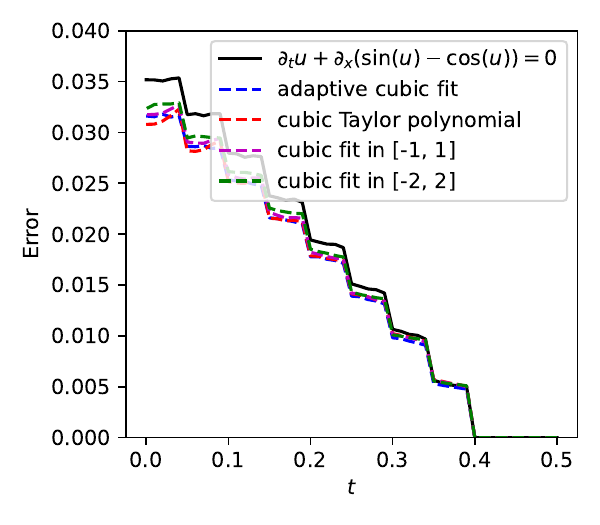}
    \caption{}
    \label{fig:results_new_cubicfit_sin_3}
    \end{subfigure}
        \begin{subfigure}{0.24\textwidth}
    \centering
    \includegraphics[width=\linewidth]{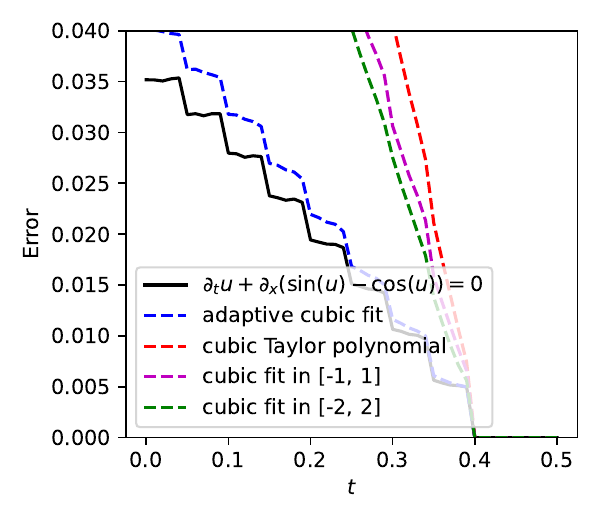}
    \caption{}
    \label{fig:results_new_cubicfit_sin_4}
    \end{subfigure}
    \caption{Generalization for $\partial_t u + \partial_x(\sin(u)-\cos(u)) = 0$. (a) the error of forward prediction for different equations. (b) The error of forward prediction with prompted examples coming from different equations. (c) the error of reverse prediction for different equations. (d) The error of reverse prediction with prompted examples coming from different equations.}
    \label{fig:results_new_cubicfit_sin}
\end{figure}

\begin{figure}[htbp]
    \centering
    \begin{subfigure}{0.24\textwidth}
    \centering
    \includegraphics[width=\linewidth]{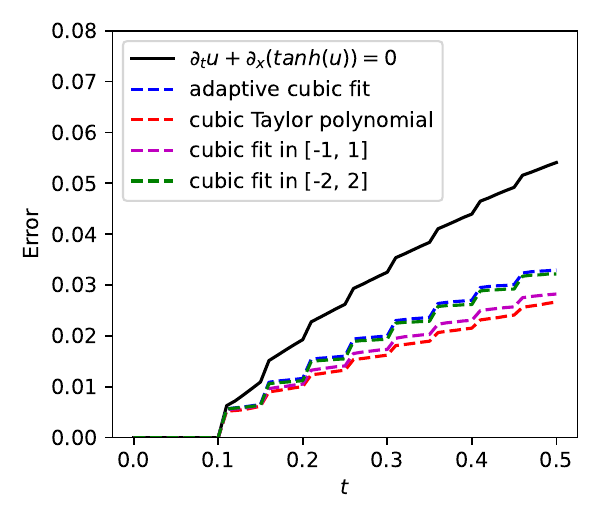}
    \caption{}
    \label{fig:results_new_cubicfit_tanh_1}
    \end{subfigure}
    \begin{subfigure}{0.24\textwidth}
    \centering
    \includegraphics[width=\linewidth]{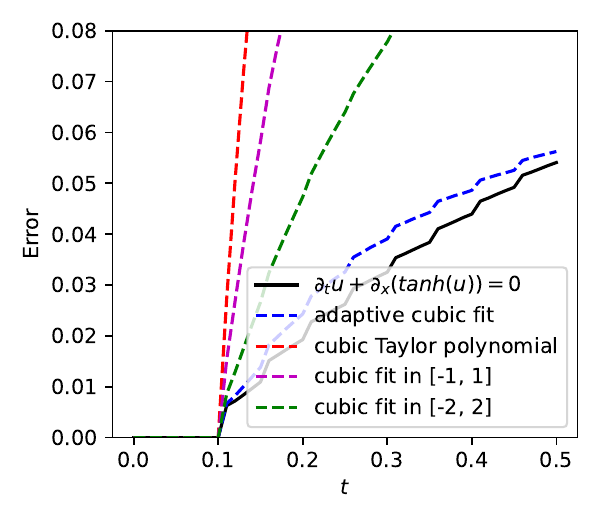}
    \caption{}
    \label{fig:results_new_cubicfit_tanh_2}
    \end{subfigure}
    \begin{subfigure}{0.24\textwidth}
    \centering
    \includegraphics[width=\linewidth]{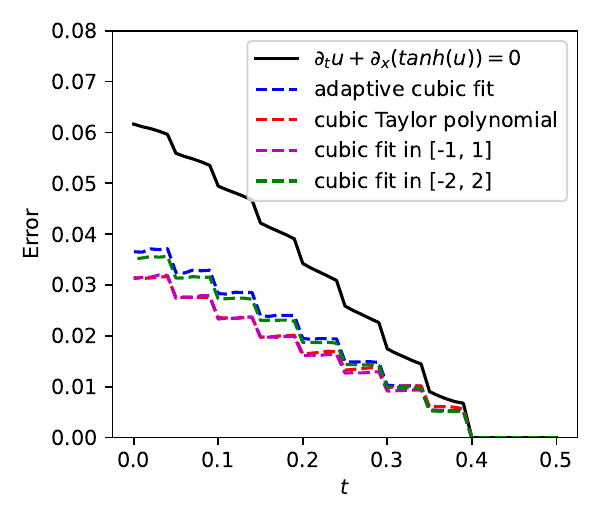}
    \caption{}
    \label{fig:results_new_cubicfit_tanh_3}
    \end{subfigure}
        \begin{subfigure}{0.24\textwidth}
    \centering
    \includegraphics[width=\linewidth]{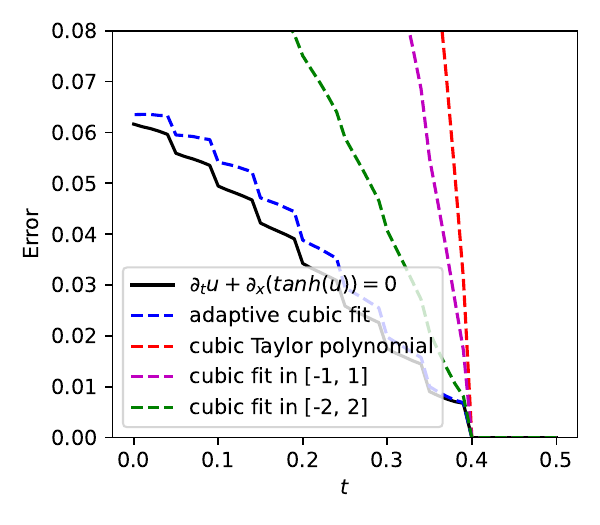}
    \caption{}
    \label{fig:results_new_cubicfit_tanh_4}
    \end{subfigure}
    \caption{Generalization for $\partial_t u + \partial_x(\tanh(u)) = 0$. The meaning of subfigures are the same as in Figure~\ref{fig:results_new_cubicfit_sin}.}
    \label{fig:results_new_cubicfit_tanh}
\end{figure}

In Figure~\ref{fig:results_new_cubicfit_sin} we show the errors of forward and reverse predictions w.r.t. time. These errors are averaged over 512 instances, with the initial conditions sampled from the same stochastic process as in training (with different random seeds).

There are two key observations:
\begin{enumerate}
    \item For Comparison 1, the error for the new equation is higher than those for ``similar'' cubic flux functions, but still within a reasonable range. This is expected since the new equation is out of the training distribution.
    \item For Comparison 2, when the prompted examples are generated from ``similar'' cubic flux functions, the errors are higher than that with ``correct'' examples. This shows that the ICON model didn't simply memorize the cubic flux functions and approximate the new flux with the closest cubic function (at least not in a trivial way). Instead, it is able to generalize to more general forms of flux functions.
\end{enumerate}

In Figure~\ref{fig:results_new_cubicfit_tanh}, we also show the results for $f(u) = \tanh(u)$, with ``similar'' cubic functions illustrated in Figure~\ref{fig:results_new_fit_2}. The observations are consistent with those for $f(u) = \sin(u)-\cos(u)$.

\subsection{Change of Variables}\label{sec:results_variable}

\begin{figure}[htbp]
    \centering
    \begin{subfigure}{0.3\textwidth}
        \centering
        \includegraphics[width=\linewidth]{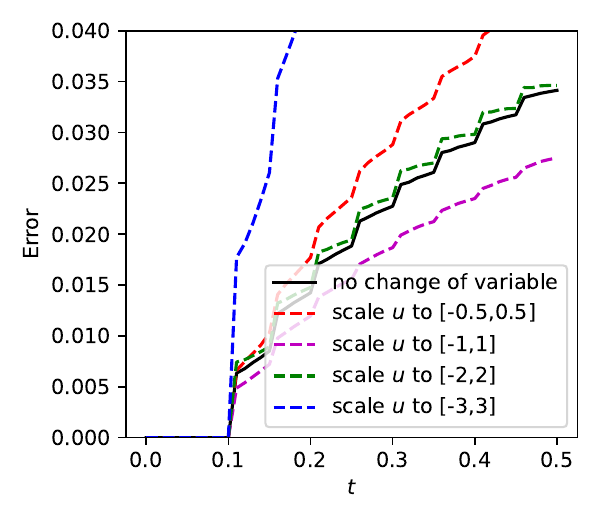}
        \caption{}
        \label{fig:results_new_variable_1}
        \end{subfigure}
        \begin{subfigure}{0.3\textwidth}
        \centering
        \includegraphics[width=\linewidth]{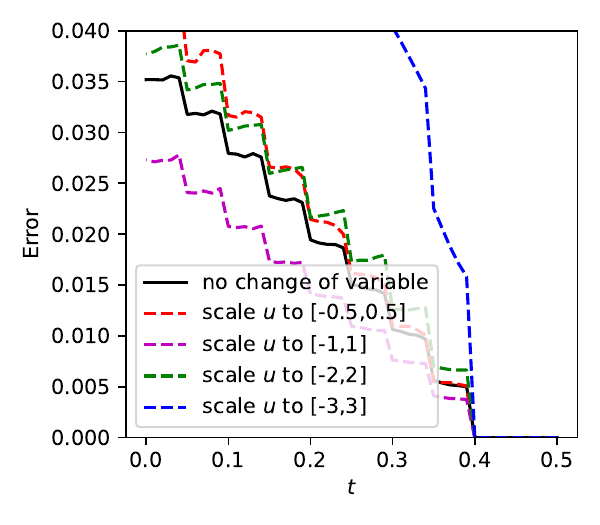}
        \caption{}
        \label{fig:results_new_variable_2}
        \end{subfigure}
    \caption{The error of prediction for $\partial_t u + \partial_x(\sin(u)-\cos(u)) = 0$, with and without change of variables. (a) the error of forward prediction. (b) the error of reverse prediction. The error is the lowest when scaling $u$ to $[-1,1]$.}
    \label{fig:results_new_variable}
\end{figure}

In this section, we study the change of variables, focusing on the example equation $\partial_t u + \partial_x(\sin(u)-\cos(u)) = 0$.

We apply the affine transformation $v = (u-\beta)/\alpha$, so that $\partial_t v +  \partial_x f(\alpha v + \beta)/\alpha = 0$. Let $\beta= (u_\text{max} + u_\text{min}) / 2$, $\alpha = (u_\text{max} - u_\text{min}) / (2 r)$, so that if $u \in [u_\text{min}, u_\text{max}]$, then $v \in [-r, r]$. Here we set $u_\text{min}$ and $u_\text{max}$ as the minimum and maximum values of $u$ in the condition-QoI examples as well as the question condition. We feed $v$ to the ICON model to make predictions, and then apply the inverse transformation $u = \alpha v + \beta$ to recover the predictions in the original variable $u$. The other setups are the same as in Section~\ref{sec:generalization}.

The errors for forward and reverse predictions are shown in Figure~\ref{fig:results_new_variable}. We can see that when $r = 1$, i.e. scaling $u$ to $[-1,1]$, the errors are the lowest. Recall the initial condition is sampled from the Gaussian process with mean 0 and variance 1, it is reasonable that $r = 1$ works better than ``no change of variables'', or $r = 2, 3$. The errors for $r = 0.5$ are higher, which can be attributed to that the prediction errors are amplified when transforming back to the original variables.

\subsection{Varying Strides}\label{sec:results_time}

\begin{figure}[htbp]
    \centering
  \begin{subfigure}{0.24\textwidth}
    \centering
    \includegraphics[width=\linewidth]{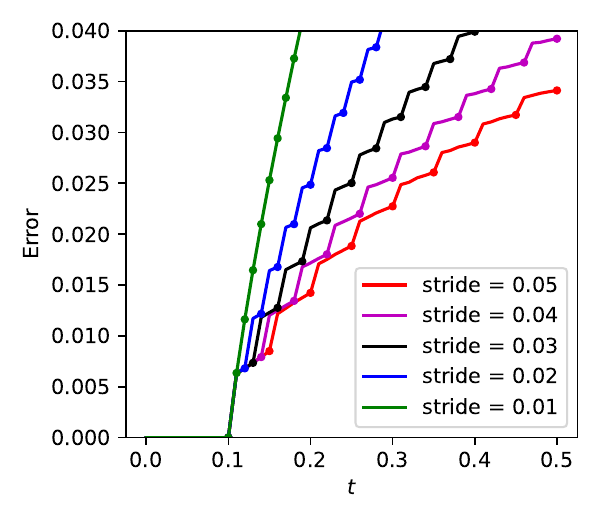}
    \caption{}
    \label{fig:results_new_time_1}
    \end{subfigure}
    \begin{subfigure}{0.24\textwidth}
    \centering
    \includegraphics[width=\linewidth]{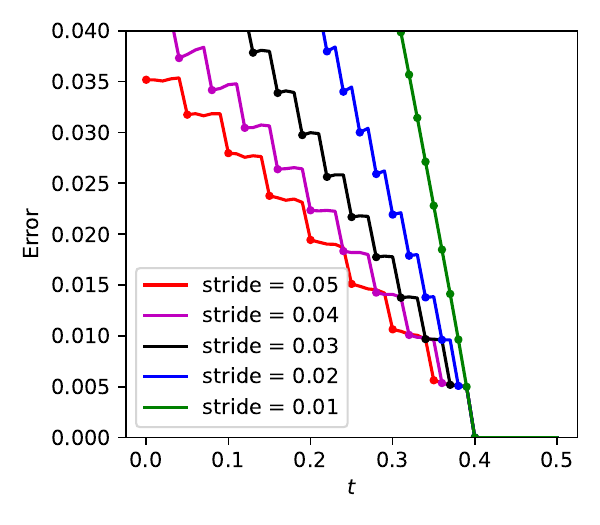}
    \caption{}
    \label{fig:results_new_time_2}
    \end{subfigure}
    \begin{subfigure}{0.24\textwidth}
    \centering
    \includegraphics[width=\linewidth]{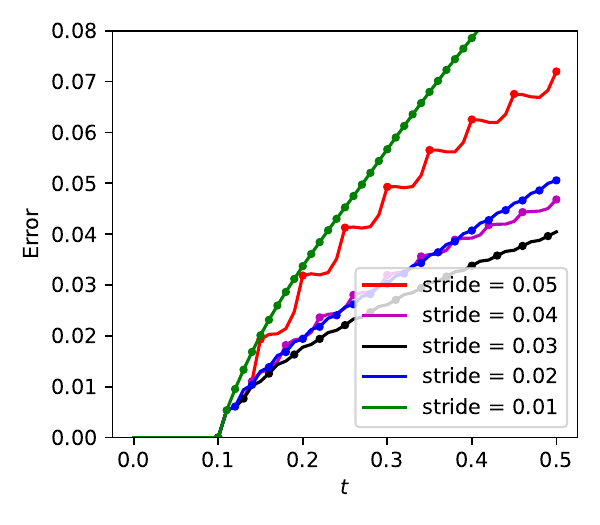}
    \caption{}
    \label{fig:results_new_time_3}
    \end{subfigure}
        \begin{subfigure}{0.24\textwidth}
    \centering
    \includegraphics[width=\linewidth]{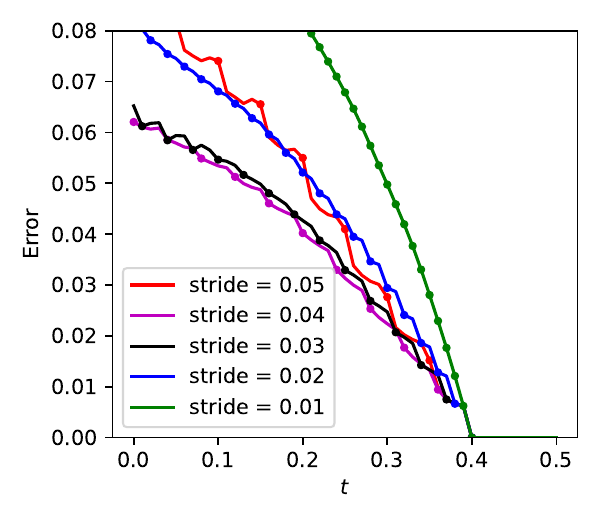}
    \caption{}
    \label{fig:results_new_time_4}
    \end{subfigure}
    \caption{The error of prediction with different maximum strides. (a,b) the error of prediction for $\partial_t u + \partial_x(\sin(u)-\cos(u)) = 0$. (c,d) the error of prediction for $\partial_t u + \partial_x(3\sin(u)-3\cos(u)) = 0$. (a,c) the error of forward prediction. (b,d) the error of reverse prediction. 
    % For $f = \sin(u)-\cos(u)$, the error increases as the maximum stride decreases from 0.05 to 0.01 due to error accumulation. For $f = 3\sin(u)-3\cos(u)$, the error decreases as the maximum stride increases from 0.01 to 0.03 due to error accumulation, and then increases as the maximum stride continues to increase from 0.03 to 0.05 due to the out-of-distribution effect. 
    }
    \label{fig:results_new_time}
\end{figure}

In this section, we study the effect of varying maximum stride $S$ for recursive predictions. We consider the equation $\partial_t u + \partial_x(\sin(u)-\cos(u)) = 0$, and  $\partial_t u + \partial_x(3\sin(u)-3\cos(u)) = 0$, and make predictions with $S = 0.01, 0.02,..., 0.05$. The other setups are the same as in Section~\ref{sec:generalization}. The results are shown in Figure~\ref{fig:results_new_time}. 

For $f = \sin(u)-\cos(u)$, the error increases as the maximum stride decreases from 0.05 to 0.01. The forward operators $\mathcal{F}_{\sin(u)-\cos(u),0.1 k}$ are equivalent to $\mathcal{F}_{k\sin(u)-k\cos(u),0.1}$ for $k = 0.5,0.4,...,0.1$. Since $\sin(u)-\cos(u)\approx -1/6 u^3 + 1/2u^2 + u + const$, these operators are in the training range where the cubic polynomial coefficients are within $[-1,1]^3$. The difference in the performance can be attributed to the error accumulation in recursive predictions, with a larger maximum stride leading to a smaller error due to less recursive steps.

For $f = 3\sin(u)-3\cos(u)$, the forward operators $\mathcal{F}_{3\sin(u)-3\cos(u),0.1 k}$ are equivalent to $\mathcal{F}_{3k\sin(u)-3k\cos(u),0.1}$ for $k = 0.5,0.4,...,0.1$. One can see that the operators are in the training range for $k = 0.1,0.2,0.3$, while out of the training range for $k = 0.4,0.5$. Therefore, the error decreases as the maximum stride increases from 0.01 to 0.03 due to error accumulation, and then increases as the maximum stride continues to increase from 0.03 to 0.05 due to the out-of-distribution effect.

\section{Summary}\label{sec:summary}
In this paper, we present a detailed methodology of ICON for PDEs with temporal evolution, using 1D scalar nonlinear conservation laws as an example. By designing the data prompts appropriately, we can use a single ICON model to make forward and reverse predictions for different equations with different strides. We show that an ICON model trained on conservation laws with cubic flux functions can generalize well to some other flux functions of more general forms, without fine-tuning. We also show how to broaden the range of problems that ICON can address, by transforming functions and equations to ICON's capability scope via change of variables and proper strides. We believe that the progress in this paper is a significant step towards the goal of training a foundation model for scientific machine learning under the in-context operator learning framework.

There are still several issues that require further investigation. Firstly, the training scheme with consistency loss remains to be improved. Secondly, the current recursive schemes with varying strides and the change of variables are still relatively simple. Thirdly, the study in this paper is limited to 1D scalar conservation laws, mainly due to limited computational resources. We leave these issues to future work.

\section*{Acknowledgement}
We acknowledge valuable discussions with Prof. Chi-Wang Shu, Dr. Tingwei Meng, Dr. Siting Liu, and Yuxuan Liu. S. Osher is partially funded by AFOSR MURI FA9550-18-502 and ONR N00014-20-1-2787.
We would like to express our gratitude to ChatGPT for enhancing the wording during the paper writing phase.

\bibliographystyle{unsrt}
\bibliography{biblist}

\newpage

\end{document}